# Representing Objects, Relations, and Sequences

**Stephen I. Gallant and T. Wendy Okaywe**

**Pitney Bowes**
**MultiModel Research**

**MultiModel Research**

**sgallant@mmres.com, wokaywe@mmres.com**

**February 2, 2013**

## Abstract

Vector Symbolic Architectures (VSAs) are high-dimensional vector representations of objects (eg., words, image parts), relations (eg., sentence structures), and sequences for use with machine learning algorithms. They consist of a vector addition operator for representing a collection of unordered objects, a Binding operator for associating groups of objects, and a methodology for encoding complex structures.

We first develop *Constraints* that machine learning imposes upon VSAs: for example, similar structures must be represented by similar vectors. The constraints suggest that current VSAs should represent phrases ("The smart Brazilian girl") by binding sums of terms, in addition to simply binding the terms directly.

We show that matrix multiplication can be used as the binding operator for a VSA, and that matrix elements can be chosen at random. A consequence for living systems is that binding is mathematically possible without the need to specify, in advance, precise neuron-to-neuron connection properties for large numbers of synapses.

A VSA that incorporates these ideas, MBAT (Matrix Binding of Additive Terms), is described that satisfies all Constraints.

With respect to machine learning, for some types of problems appropriate VSA representations permit us to *prove* learnability, rather than relying on *simulations*. We also propose dividing machine (and neural) learning and representation into three Stages, with differing roles for learning in each stage.

For neural modeling, we give "representational reasons" for nervous systems to have many recurrent connections, as well as for the importance of phrases in language processing.

Sizing simulations and analyses suggest that VSAs in general, and MBAT in particular, are ready for real-world applications.

# 1. Introduction

Representation is an important topic in its own right.

Perhaps the most successful representation ever invented (other than writing itself!) is the decimal representations of integers, a great advance over counting by simple hash marks. Decimal representations illustrate that a good representation can help us calculate more quickly, by orders of magnitude, and thereby enable computations that would otherwise be impossibly difficult. This is precisely our goal here. We want a representation of objects (for example, words), multiple relations of those objects (eg., assorted syntactic and semantic information), and sequences (eg., sentences), that is hospitable to machine learning. For another example in the computer vision domain, we are interested in representing image parts, relations among those parts, and sequences of (sub) image presentations (eg., from eye saccades).

We seek a representation that permits us to use standard machine learning techniques (eg., neural networks, perceptron learning, regression) to *simultaneously* learn mappings of objects, relations, and sequences. Moreover, we want to use standard algorithms "out of the box" on vector inputs, without the need for constructing a separate learning architecture for each task.

This would open the possibility of "higher order holistic modeling", where the predicted output encodes objects simultaneously with their structure, and where the structure can be more complex than selection from a small set of options. For example, we want to be able to predict full parse trees in one shot, rather than word-for-word part-of-speech tags. Ultimately we would like to predict a translated or summarized sentence, or a transformed image representation.

These are longer term goals; a more immediate motivation for developing such representations is to facilitate the use of machine learning when starting from the outputs of Structured Classification approaches. For example, Collobert et al. [2011] produce a system that outputs structure information (part of speech, chunks, semantic roles) for each word in a sentence. We want to be able to cleanly incorporate these outputs into a fixed-length vector representing the entire sentence, for use with follow-on machine learning.

To address these goals, since the 1990s a number of investigators have worked on incorporating structure into high-dimensional, distributed vector representations. (A *distributed vector* represents objects or other information by patterns over the entire vector.) Following Levy & Gayler [2008], we'll refer to these architectures as *Vector Symbolic Architectures* (VSAs).

The desire to use machine learning techniques places a number of Constraints on *Representation*. Inputs and outputs to standard machine learning algorithms are most conveniently expressed as fixed-length vectors, ie., vectors having a pre-specified number of components. Thus we cannot directly apply neural networks to sentences, because sentences have no fixed and bounded length (and also because they possess important structure that is not immediate from the string of letters in a sentence).

Here we will focus on developing a VSA that simultaneously represents multiple objects, multiple versions of relations among those objects, and sequences of such objects/relations using a *single* fixed-length vector, in a way that satisfies the representational constraints. We name the VSA we develop *Matrix Binding of Additive Terms* or MBAT.

## Vector Symbolic Architectures

To help with basic intuition for VSAs, consider Table 1, where five terms ("*smart*", "*girl*", "*saw*", "*grey*", "*elephant*") are shown with their corresponding vectors ($\mathbf{V}^{\mathbf{smart}}$, $\mathbf{V}^{\mathbf{girl}}$, etc).

Notationally, we represent all matrices by **M**, and vectors by other capital letters, such as **V**,**W**. We will also follow the standard convention of representing the vector for a term ($\mathbf{V}^{\mathbf{smart}}$) by just the term (**smart**) where the context is clear.

Vectors for words:

| smart | girl | saw | gray | elephant |
|---|---|---|---|---|
| -1 | 1 | -1 | 1 | 1 |
| 1 | 1 | -1 | -1 | -1 |
| 1 | 1 | 1 | -1 | 1 |
| -1 | -1 | 1 | -1 | -1 |
| -1 | -1 | 1 | -1 | -1 |
| -1 | 1 | -1 | -1 | 1 |
| 1 | -1 | 1 | -1 | -1 |
| -1 | -1 | 1 | 1 | 1 |
| -1 | -1 | 1 | 1 | -1 |
| 1 | -1 | -1 | -1 | -1 |

| V = smart + girl |
|---|
| 0 |
| 2 |
| 2 |
| -2 |
| -2 |
| 0 |
| 0 |
| -2 |
| -2 |
| 0 |

Dot products with V:

| smart | girl | saw | gray | elephant |
|---|---|---|---|---|
| **12** | **12** | **-8** | **-4** | **4** |

**Table 1**: Computational example with 10-dimensional vectors. This illustrates the sum of two vectors, and the process for recognizing individual constituents from a sum using the dot product.

Table 1 suggests that we have a way of recognizing individual constituents of a vector sum using dot products (vector inner products). This will be formalized below in Section 3.

Vector Symbolic Architectures trace their origins to Smolensky's [1990] tensor product models, but avoid the exponential growth in vector size of those models. VSAs include Kanerva's *Binary Spatter Codes* (BSC) [1994, 1997], Plate's *Holographic Reduced Representations* (HRR) [1992, 2003], Rachkovskij and Kussul's *Context Dependent Thinning* (CDT) [2001], and Gayler's *Multiply-Add-Permute* coding (MAP) [1998].

Vector Symbolic Architectures can be characterized along five defining characteristics:

- Components of vectors are either binary (BSC), sparse binary (CDT), “bi-polar” (+1/-1) (MAP objects), continuous (HRR and MAP sums), or complex (HRR).

- Addition of vectors (also referred to as “*bundling*”) represents collections of (simple or complex) objects, but without any structure among the summed terms. If objects represent words, their addition gives an unordered “bag of words.” Operators used for addition include normal vector addition as in Table 1 (HRR, MAP), and addition followed by conversion to binary components according to thresholds (BSC).

- Binding of vectors is used to group objects, and can also be used for ordering them. Binding operators include Exclusive-OR or parity (BSC) and component-wise multiplication (MAP).

  A particularly important binding method is circular convolution (HRR). Letting **D** be vector dimensionality, the circular convolution of two vectors, $\mathbf{V} = \mathbf{X} * \mathbf{Y}$, is defined by

  $$\mathbf{V_j} = \sum_{(k = 0, \ldots D-1)} \mathbf{X}_k \, \mathbf{Y}_{j-k}$$

  where the subscript calculation is taken mod D. In other words, reverse the numbering of **Y**’s indices, and now each component of the result $\mathbf{V}_j$ is just the dot product of **X** and (reverse numbered) **Y**, where **Y** is first rotated j positions prior to taking the dot product.

  Binding is commutative with respect to its two operands, and VSAs typically include an inverse operation for recovering one operand if the other is known. Inverses can be mathematically exact inverses (BSC, MAP) or have mean-0 noise added to the result (HRR), in which case a “cleanup step” is required to find the unknown element. *Cleanup* consists of finding the closest resulting vector using dot products with all vectors, or making use of Auto-associative memories [Kohonen 1977, Anderson et al. 1977].

  For CDT binding, sparse binary vectors (representing objects or sub-structures) are first OR’ed together forming vector **V**. Then **V** is AND’ed with the union of a fixed number of permutations of **V** to control the expected number of 1s in the final vector. A separate addition operator is not needed for CDT.

- Quoting applied to binding produces unique binding operators in order to differentiate among groups joined by binding. This typically involves a random permutation of vector elements to represent, for example, two different subject phrases in a single sentence. (“*The smart girl* *and* *the grey elephant* *went for a walk.*”)

- Complex Structure Methodology represents complex relations among objects, such as nested sub-clauses in a sentence. For VSAs, this consists of binding (and quoting) to get sub-objects bound together, and addition to represent unordered collections of bound sub-objects. For example, let us suppose each of the roles actor, verb, and object have their own vectors, as well as objects *Mary*, *loves*, and *pizza*. Then using * to denote binding in VSAs, we might represent “*Mary loves pizza*” by the vector

  **(actor * Mary) + (verb * loves) + (object * pizza)**.

  This permits extraction of, say, the actor (Mary) by binding the final sum with the inverse of actor, and following with a “cleanup” step.

For MBAT, we will be presenting a different, unary binding operator, and a different complex structure methodology that emphasizes additive "phrases".

### Organization

This paper is organized as follows. We first propose in Section 2 a collection of necessary *Constraints* for representing structured objects for use by standard machine learning algorithms. We then (Section 3) describe the MBAT architecture that encodes objects, structures and sequences into a single distributed vector. In Section 4, we examine the role (and advisability) of machine learning during three information processing stages: Preprocessing, Representation Generation, and Output Computation. We also see how for many tasks, with a suitable representation we can *prove* learnability for standard machine learning algorithms, rather than rely upon *simulations*.

Section 5 looks at capacity, namely the required dimensionality for vectors. Both analytic estimates and simulation results are presented. Section 6 re-examines the Constraints with respect to the MBAT architecture. Section 7 reviews prior research. Section 8 (Discussion) revisits VSAs with respect to Complex Structure Methodology, and suggests applications for MBAT in computational linguistics, computer vision and modeling neural information processing. The Appendix develops estimates for required dimensionality of vectors.

## 2. Requirements for a Good Representation of Objects, Relations and Sequences

We want a representation of structured objects, such as sentences or images, that is directly suitable for machine learning, without the need for constructing special-case learning algorithms. There are several requirements that must be met:

> **Constraint 1**: **Fixed Length Vector**. Most standard machine learning approaches take inputs that are vectors of some pre-specified length. Thus if we want a way to simultaneously learn mappings of objects and structures, we need a way to represent many different objects, and structures of those objects, simultaneously, in a single vector with pre-specified length, eg., 1,000 components. (For simplicity and concreteness, we refer here to a 1,000-dimensional system. However, a practical system may require a different dimensionality, either larger for increased capacity or smaller for increased speed of computations. Section 5 and the Appendix explore dimensionality requirements.)
>
> **Constraint 2**: **Distributed Representations**. We need to represent hundreds of thousands of objects involved in an exponentially larger number of representations, so only one bit or vector component per object will not supply sufficient capacity. Therefore, we need to use a *distributed representation* for the vector, where information is stored in patterns, and where an individual component gives little, if any, information.
>
> To take a specific example in natural language, we might represent a word as a 1,000 dimensional vector, whose components are randomly generated choices of -1 and +1 (as in Table 1). Then we can represent a sentence as the (single vector!) sum of the vectors for words in the sentence. We'll examine disadvantages of this representation in Constraint 4, but the sum of vectors gives one way to represent a variable length sentence as a single, distributed, fixed-length vector.

It is amusing to note that the usual computer representation of a sentence as a text string qualifies as a distributed representation! Any individual letter gives little or no information; only the larger pattern of letters gives information. Similarly, an image bit map is also a distributed representation. These representations have a minor problem in that they are not fixed length, but they also have a major "continuity" problem, as discussed below in Constraint 4.

**Constraint 3: A Complex Structure Methodology for Representing Objects and Structure Information Within the Distributed Vector.** For many natural-language tasks, we clearly must take the syntactical structure into account. Here we encounter the "Binding Encoding Problem" in Cognitive Science and Artificial Intelligence surveyed by Treisman [1999]: for the word pair "*smart girl*", we need to represent that "*smart*" refers to "*girl*", and not some other word in the sentence. More generally, we need to be able to represent full parse information (or relations among image features) in a single distributed vector. This includes representing sub-clauses in sentences and representing parts of images with associated features (eg., color, location, motion).

Conversely, given a vector, we need to be able to recognize objects or structures encoded in the vector.

**Constraint 4: Map Similar Objects and Structures to Similar Representations**. For learning algorithms to be able to generalize, it is necessary that similar objects and structures be represented by similar vectors. This is a continuity property for maps from objects and their structures to their representations.

On the representation side, vector similarity is readily defined by Euclidean distance between vectors. Two vectors are similar if (after normalization) they are close in Euclidean distance or, equivalently, if they have a significantly greater dot product than the dot product for two randomly chosen vectors.

Starting with *object similarity*, we need to represent similar objects by similar vectors. For example, we want the vector for "*smart*" to be similar to the vector for "*intelligent*".

Turning to *structure representations*, we also need to represent similar structures by similar vectors. For example, we want all of the following to have similar vectors (to varying degrees):

- "*The smart girl saw the gray elephant*"
- "*The gray elephant was seen by the smart girl*"
- "*The smart girl I just met saw the young gray elephant eating peanuts*"

(The second case might also be considered as having a different *structure*, but similar *meaning*.)

For images, we want to be able to replace similar image parts, or include additional features and structure information for an image, and have the new image vector similar to the original image.

This Similarity Constraint is where character strings and bit maps fail as vector representations. For a string, if we add a space, or change to a similar word with a different number of letters, or switch active/passive voice, then the vector of letters changes drastically (as measured by vector Euclidean distance). Similarly, adding a row of pixels to an image can make a drastic difference in bit map vector similarity.

**Constraint 5:** **Sequences**. We need to represent sequences of objects and relations. For example, we want to represent a group of sentences, as well as images derived from a sequence of eye saccades. This requirement for sequences is especially strong for spoken language, where even a single two-syllable word like "*baby*" does not hit our auditory system all at once, but rather as a sequence of sounds.

Thus, ultimately, we need to represent, over time, sequences of objects and relations: phonemes, words, sentences, images, or sensory inputs.

**Constraint 6:** **Efficient Encoding into the Representation**. If we want to be able to encode, say, 100,000 sentences as 100,000 vectors, we need the mapping computation from each sentence to its representation vector to be roughly linear in the length of the sentence (or number of objects and relations for the sentence). Methods that require a machine learning pass over all 100,000 sentences to represent one of the sentences, or that require $n^2$ computation to represent n sets of objects and structures, would seem to be impractical. Similarly, we can't practically use a representation method that, when presented with a new object, requires re-computation of the representations for all previously seen objects.

**Constraint 7:** **Neural Plausibility**. Although not required for computational applications in language, vision, etc., we are nonetheless interested in representations that can serve as abstract models that capture important representational functionality in living systems.

To summarize, we have listed six "must have" Constraints, along with one final "nice to have" Constraint for representing objects, structures, and sequences so that we can use machine learning algorithms (and their extensive mathematical theory) "out of the box," without constructing special case learning algorithms.

# 3. Representing Objects, Relations and Sequences Using a Single Distributed Vector

We now define MBAT, a Vector Symbolic Architecture, and show how it represents objects, relations and sequences by a single, distributed, fixed-length vector, while satisfying previously described Constraints.

We employ two vector operations: *Addition* (+) and *Binding* (#), as well as a *Complex Structure Methodology* of binding additive phrases, as described in the following Sub-Sections.

## 3a. Vector Addition (+) and Additive Phrases

The familiar vector addition operator is sufficient to encode an unordered set of vectors as a single vector of the same dimension as its constituent vectors. For example, in previous work we encoded a document as the sum of its constituent term vectors, and used this document vector for Information Retrieval purposes [Caid et al. 1995]. The key property of vector addition, illustrated in Table 1, is:

### ***Property 1: <u>Addition Preserves Recognition</u>***

This property is non-intuitive. For example, with scalars if we know that six positive and negative integers added together sum to 143, we cannot say whether one of those numbers was 17.

By contrast, as in Table 1, suppose we add together six 1,000 dimensional vectors with random +1/-1 components representing words,

$$\mathbf{V}^{\mathbf{Sum}} = \mathbf{V}^{\mathbf{1}} + \mathbf{V}^{2} + \ldots + \mathbf{V}^{\mathbf{6}}.$$

Let us denote the vector for the term "*girl*" by $\mathbf{V}^{\mathbf{girl}}$**.** Now we can be highly certain whether $\mathbf{V}^{\mathbf{girl}}$ was one of the six. We simply compute the inner product (dot product)

$$\mathbf{x} = \mathbf{V}^{\mathbf{girl}} \bullet \mathbf{V}^{\mathbf{Sum}} = \sum \mathbf{V}_{\mathbf{i}}^{\mathbf{girl}} \ \mathbf{V}_{\mathbf{i}}^{\mathbf{Sum}}$$

and if **x** is near 1,000 the answer is "yes", while if **x** is near 0 then the answer is "no".

**<u>Proof</u>:** If $\mathbf{V}^{\mathbf{girl}}$ is one of the six vectors, say $\mathbf{V}^{\mathbf{1}}$, then

$$\mathbf{V}^{\mathbf{girl}} \bullet \mathbf{V}^{\mathbf{Sum}} = \mathbf{V}^{\mathbf{girl}} \bullet (\mathbf{V}^{\mathbf{girl}} + \mathbf{V}^{2} + \ldots + \mathbf{V}^{6})$$

$$= \mathbf{V}^{\mathbf{girl}} \bullet \mathbf{V}^{\mathbf{girl}} + \mathbf{V}^{\mathbf{girl}} \bullet (\mathbf{V}^{2} + \ldots + \mathbf{V}^{6})$$

$$= \mathbf{1{,}000} + \langle \textbf{mean 0 noise} \rangle$$

Similarly, if $\mathbf{V}^{\mathbf{girl}}$ is not one of the six vectors, then

$$\mathbf{V}^{\mathbf{girl}} \bullet \mathbf{V}^{\mathbf{Sum}} = \langle \textbf{mean 0 noise} \rangle$$

This completes the proof except for one small point: we have to verify that the standard deviation of the **<mean 0 noise>** term does not grow as fast as the vector dimension (here 1,000), or else the two dot products could become overwhelmed by noise, and indistinguishable for practical purposes. The Appendix shows that the standard deviation of the noise grows by the square root of the vector dimension, completing the proof. ∎

The Addition Property of high-dimensional vectors gets us part of the way to a good distributed representation for a collection of objects. For example, we can represent a sentence (or a document or a phrase) by a single (normalized) 1,000 dimensional vector consisting of the sum of the individual word vectors. Then we can compute the Euclidean distance between vectors to find, for example, documents with vectors most similar to a query vector. This was the approach for our previous document retrieval efforts. However, we still need to represent structure among objects.

## 3b. The Binding Operator (#)

For both language and vision, relying solely on vector addition is not sufficient. Due to the commutativity of vector addition, multiple phrases such as in "*The smart girl saw the gray elephant*" will have exactly the same vector sum as "*The smart elephant saw the gray girl*" or even "*elephant girl gray saw smart the the*". In other words, vector addition gives us the "bag of words" used to create the sum, but no other structure information.

Here we run into the classic "Binding Encoding Problem" in Cognitive Science and Artificial Intelligence, surveyed by Treisman [1999]. We need some way to bind "*gray*" to "*elephant*" and not to "*girl*" or to any other word, while retaining a distributed representation. More generally, we need the ability to represent a parse tree for a sentence, yet without abandoning distributed representations.

### Phrases

It is first helpful to formalize the definition of *phrase* with respect to representations. We define a *phrase* as a set of items that can have their order changed without making the representation unusable. Phrases *loosely* correspond to language phrases, such as noun clauses and prepositional phrases, or "chunks" in computational linguistics. For example, in *"The smart Brazilian girl saw a gray elephant,"* we can reorder the leading four-word noun phrase as in *"Brazilian the girl smart saw a gray elephant,"* and still understand the sentence, even though it becomes ungrammatical.

Similarly, for machine vision, an example of a phrase would be the vectors for an associated shape, X and Y-positions, color, and motion. Again, order is not critical.

### Neural Motivation

To motivate the binding operator we propose, consider the neural information processing schematic in Figure 1.

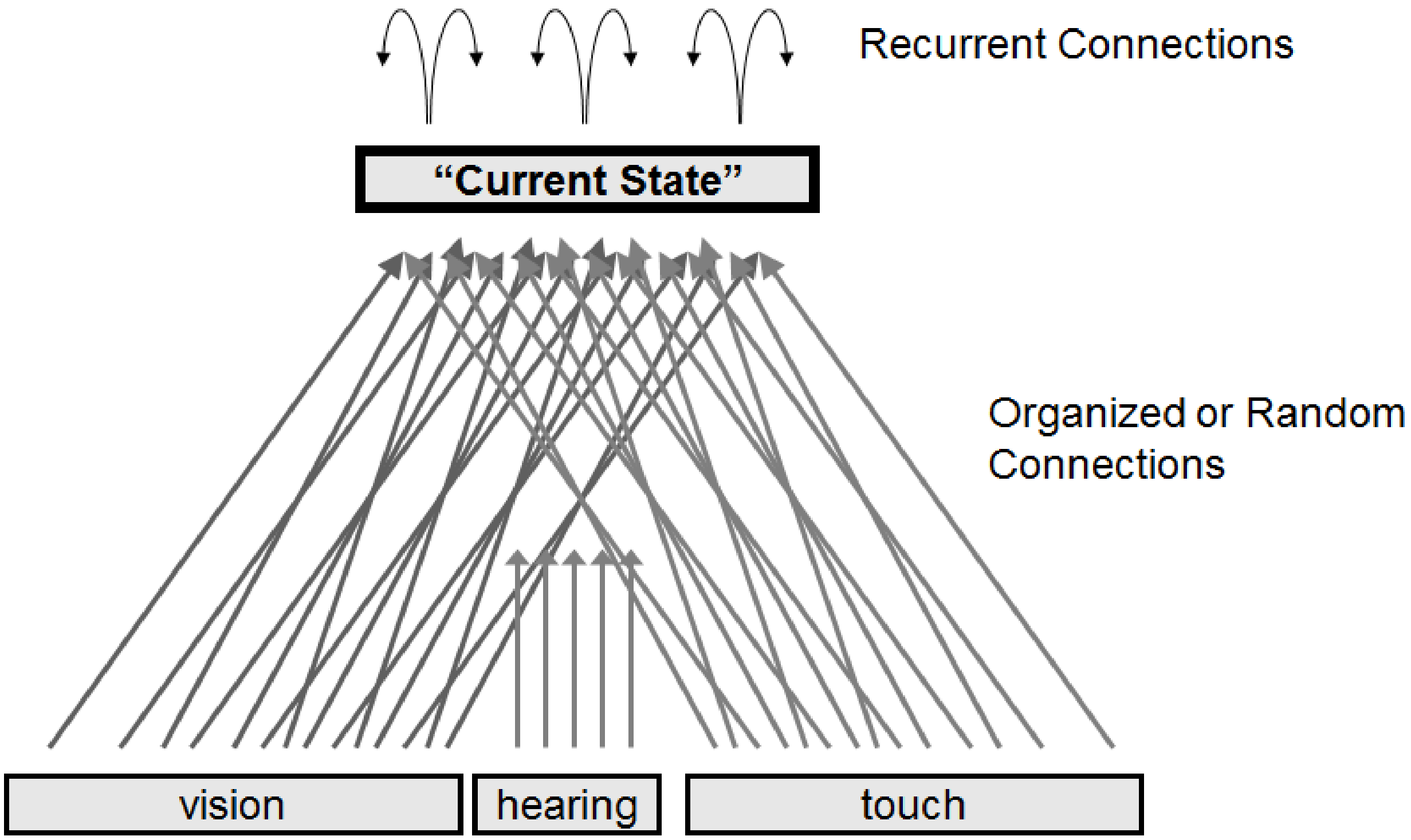

**Figure 1:** Recurrent connections give a way to bind inputs to the current and previous system states, as well as each other. We will see that these connections may be randomly generated.

Here we have inputs from various sensory subsystems: vision, hearing, and touch. The current state of the system ("neuron state") is modified by information from these inputs, as well as recurrent connections from itself.

The Figure illustrates the "Brain Binding Problem," where we need the capability of linking together diverse sensory inputs (or different neural regions), with the current state of the system. Sequences also come into play here, as when we hear "*baby*" as two phonemes over two time periods, we need to sequentially bind the inputs to recognize and represent the term "*baby*" and its associations.

For the binding task, the main thing we have to work with are the recurrent connections at the top of the Figure. (Any Theory of Neural Information Processing that does not include a major role for such recurrent connections is missing a very big elephant in the Neural Physiology room!) Moreover, we cannot make too many organizational demands upon the recurrent connections, because any complex structure needs to be passed genetically and hooked up during a noisy, messy growth phase.

So, continuing with our motivational exploration for binding, what if we take the easiest possible genetic/growth structural organization, namely *random*. Can we have the recurrent connections compute a random map and have that be of any use for binding?

## Binding Operator (#)

Returning to the mathematics, let us now define the simplest version of a unary binding operator, **#**. (Below we will also define several alternatives.)

Let **M** be a fixed square matrix of appropriate dimension for our vectors, eg., 1,000 by 1,000. We let components of **M** be randomly chosen values (eg., +1/-1).

As a point of notation, when raising a *matrix to a power*, we will always use parentheses, as in $(\mathbf{M})^3$. This distinguishes from the designation of several *different matrices*, for example $\mathbf{M}^{\text{Actor}}$ and $\mathbf{M}^{\text{Object}}$.

Now if we have a sum of vectors, $\mathbf{V}^1 + \mathbf{V}^2 + \mathbf{V}^3$, i.e., a phrase, we can bind them as part of a structure description by:

$$\#(\mathbf{V}^1 + \mathbf{V}^2 + \mathbf{V}^3) \equiv \mathbf{M}(\mathbf{V}^1 + \mathbf{V}^2 + \mathbf{V}^3).$$

(The "#" operator "pounds" vectors together.) Thus all terms in the vector of the form $(\mathbf{M})^1\,\mathbf{V}$ are differentiated from terms of the form $(\mathbf{M})^2\,\mathbf{V}$, $(\mathbf{M})^3\,\mathbf{V}$, etc. We can think of $(\mathbf{M})^i\,\mathbf{V}$ as transforming **V** into a unique "bind space" according to *i*.

With respect to complex structure methodology, in MBAT *binding operates upon additive phrases*, where the order of vectors in the phrase is not critical. Thus we bind

**#(<u>actor</u> + the + smart + Brazilian + girl)**

**≡ M (<u>actor</u> + the + smart + Brazilian + girl).**

Each term in the phrase may itself be the result of a binding operation, which allows us to represent complex structure (for example, sub-clauses of a sentence).

Some things to note:

- One of the vectors in the summed arguments can be the current state of the system, so letting

  **V(n)** be the current state at time **n**,

  we have the next state given by

  $$\mathbf{V(n+1)} = \mathbf{M}\,(\mathbf{V(n)}) + \sum \mathbf{V}^{\text{inputs}} \qquad (3.1)$$

  Note that the Binding operator in this formulation corresponds to the recurrent connections in Figure 1. $\mathbf{M}_{i,j}$ is the synapse between cell *j* and cell *i*. (Also, individual cells in the "Current State" do not need to differentiate whether inputs are coming from feed-forward sources or from recurrent connections.)

- This formula for computing the next state also gives a way to represent input *sequences*. Kanerva [2009] and Plate [2003] previously employed this technique for sequence coding, using different binding operators.

- Phrases that get bound together must be unambiguous with respect to order. Thus we can bind phrases like "*the smart girl*", where order doesn't really matter in understanding the phrase. However, we couldn't bind in one step "*the smart girl saw the grey elephant*", because we would run into the binding ambiguity of whether "*smart*" refers to "*girl*" or "*elephant*". Several binding operations would be required, as in Figure 2.

- We can make good use of Tags, represented by (random) tag vectors added to phrases, to specify additional syntactic and semantic information such as **actor** (ie., $\mathbf{V}^{\mathbf{actor}}$), **object**, **phraseHas3words**, etc.

- Computing binding (matrix multiplication) involves more work than computing circular convolution in Holographic Reduced Representations if Fast Fourier Transforms are used for HRRs [Plate 2003]. Also, Binding in MBAT requires us to make use of a different mathematical space, i.e., matrices vs. vectors-only in HRRs.

Now we can see how to unambiguously represent "*The smart girl saw the gray elephant*" in Figure 2.

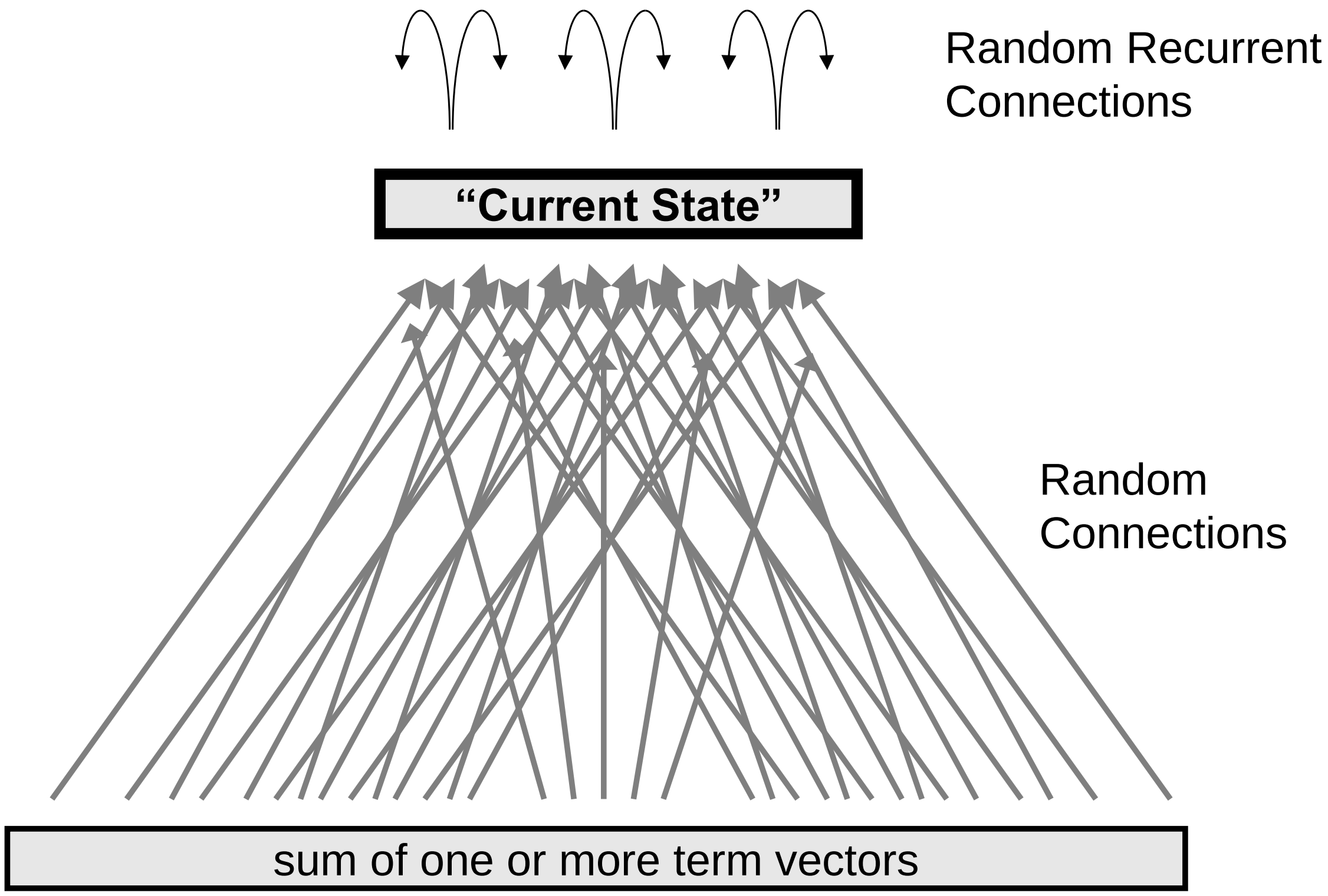

| Time | Vector |
|---|---|
| **1.** | **actor + the + smart + girl + phraseHas3words** |
| **2.** | **verb + saw + phraseHas1word** |
| **3.** | **object + the + gray + elephant + phraseHas3words** |

**Figure 2: Representing a sentence by binding a sequence of phrase inputs.** At each time step, a phrase consisting of a sum of word vectors is collected in the bottom vector. The phrase sum may also contain structure information (eg., **subject**, **passive-voice**) in the form of Tag vectors. This vector is added to the random recurrent connections from **V(n)**, to produce the next state vector, **V(n+1)**.

The resulting (single) vector, **V**, is formed from 13 object/Tag vectors:

**V = $(M)^2$ (actor + the + smart + girl + phraseHas3words) + M (verb + saw + phraseHas1word) + (object + the + gray + elephant + phraseHas3words).**

Tags such as **phraseHas3words** and **phraseHas1word**, though perhaps not biologically realistic, greatly simplify the task of decoding the vector, i.e. producing the sentence encoded in the sum.

If we need to speak the sentence, an approach to "decoding a vector" is to produce each phrase by first computing the number of words in the phrase, and then finding that many terms with the highest dot products.

As desired, "*smart*" is associated with "*girl*" in this sum of 13 vectors, because we have term $\mathbf{(M)^2\ V^{smart}}$ and $\mathbf{(M)^2\ V^{girl}}$, but elephant appears as $\mathbf{V^{elephant}}$.

We also have a recognition property for the binding operator.

***Property 2: <u>Binding (#) Preserves Recognition</u>***

Suppose we are given $\mathbf{V = \#(\ V^1 + V^2 + V^3)}$. Can we tell if $\mathbf{V^{girl}}$ is among the bound operands?

Yes, we simply look at $\mathbf{M\ V^{girl} \bullet V}$

$$\mathbf{= M\ V^{girl} \bullet (\ M\ V^1 + M\ V^2 + M\ V^3)} \qquad (3.2)$$

and the result follows similarly to Property 1 for vector sums.

∎

## 3c. Complex Structure Methodology

Figure 2 also illustrates the Complex Structure Methodology we employ in MBAT. Binding is a *unary operator* that operates upon *phrases* consisting of *bundled* (added) vectors. Each vector being summed may be an object (eg., word), or the result of another binding operation (eg., sub-clause). Thus "*the smart Brazilian girl"* is represented by

**#(<u>actor</u> + the + smart + Brazilian + girl).**

Given a vector for a complex structure, we can check whether "*girl*" appears in any of the phrases at three outermost levels by taking a dot product with the single vector[1]

$$\mathbf{[(M)^0 + (M)^1 + (M)^2]\ V^{girl}}\ . \qquad (3.3)$$

The dot product with $\mathbf{V}$ given by $\mathbf{[\ ((M)^0 + (M)^1 + (M)^2)\ V^{girl}] \bullet V}$ will be large positive only if $\mathbf{V^{girl}}$ appears in one of the phrases, i.e. as $\mathbf{V^{girl}}$, $\mathbf{(M)^1\ V^{girl}}$, or $\mathbf{(M)^2\ V^{girl}}$.

Similarly, we can determine if "*smart*" and "*girl*" appear together in any phrase in $\mathbf{V}$ by checking if

$$\mathbf{MAX_{i=0}^{2}\ [(M)^i\ (\ V^{smart} + V^{girl}) \bullet V\ ]} \qquad (3.4)$$

is sufficiently positive. Note that if "*girl*" appears without "*smart*" in a phrase, then the value above is still positive, but half of the value than when both appear in the same phrase. Also note that we cannot replace the **MAX** operator by a sum, or we run into binding ambiguity issues.[2]

Thus using *additive vector phrases*, $\mathbf{(V^1 + V^2 + V^3)}$, as operands for binding helps with subsequent recognition (and learning) of items. It also helps reduce computational demands compared to using only binding operations, because vector addition is cheaper than matrix multiplication.

---

[1] $\mathbf{(M)^0}$ is the identity matrix.
[2] It is not clear to what extent **MAX** should be considered neurally plausible.

## 3d. Variants of the Binding Operator

As with vector addition, vector binding has several important variations.

- We can define a collection of binding operators with structural significance, and give each phrase its own binding operator, such as $\mathbf{M}^{\mathbf{Actor}}$ and $\mathbf{M}^{\mathbf{Object}}$. This makes all phrases at the same level. For example,

  **V = M$^{\mathbf{Actor}}$ (the + smart + girl + <u>phraseHas3words</u>) + M$^{\mathbf{verb}}$ (saw + <u>phraseHas1word</u>) + M$^{\mathbf{Object}}$ (the + gray + elephant + <u>phraseHas3words</u>).**

- As a special case, we can also define "Two Input" binding operators. For example, if we want a binary parse tree, we can define $\#(\mathbf{V}^1, \mathbf{V}^2)$ to be $\mathbf{M}^{\mathbf{Left}}\,\mathbf{V}^1 + \mathbf{M}^{\mathbf{Right}}\,\mathbf{V}^2$, where $\mathbf{M}^{\mathbf{Left}}$ and $\mathbf{M}^{\mathbf{Right}}$ are two different fixed matrices. Note that "Two Input **#"** is non-commutative:

  $$\#(\mathbf{V}^1, \mathbf{V}^2) \neq \#(\mathbf{V}^2, \mathbf{V}^1)$$

  as required for specifying a binary tree.

- "Binary World": A most interesting variation is to replace components of

  $\#(\mathbf{V}^1 + \mathbf{V}^2 + \mathbf{V}^3)$ by "+1" if greater than or equal to 0, and "-1" if less than 0 (as in Kanerva's Binary Spatter Codes). Restricting to +1/-1 components has the advantage of playing nicely with Auto-associative learning algorithms [Kohonen 1977, Anderson et al. 1977].

  It is worth noting that we can preserve many of the benefits of *continuous* vector components (eg., for vector sums), while still restricting all vector components to +1/-1. We take a group of vector components computed from (approximately) the same connections and employ different thresholds, obtaining a binary representation for a continuous sum. For example, we can replace the first continuous component, $\mathbf{V}_1$, of an input by the *group* of binary components

  $$\mathbf{V}_{1a} \equiv +1 \text{ if } (\mathbf{V}_1 \geq 37)\text{; else } -1$$

  $$\mathbf{V}_{1b} \equiv +1 \text{ if } (\mathbf{V}_1 \geq 5)\text{ ; else } -1$$

  $$\mathbf{V}_{1c} \equiv +1 \text{ if } (\mathbf{V}_1 \geq 19)\text{ ; else } -1$$

  … .

- Permutation Matrices: It is possible to use a permutation (random or not) for the binding matrix, as permutations are maximally sparse and easy to invert. However, an advantage of using matrices with many non-zero elements is that they can boost the representational dimensionality of isolated inputs. For example, suppose the goal is to learn Exclusive-OR (XOR) calculated on components 1 and 2 (and ignoring other components). A random permutation maps the two inputs to two different components but retains the same dimensionality, so that the probability of the resulting representation being linearly separable remains at 0. By contrast, in a "binary world" architecture with -1/+1 components, when a non-sparse random matrix is applied to inputs followed by a thresholding step, components 1 and 2 are spread non-linearly among many components. This increases the effective dimensionality of the representation [Gallant & Smith 1987], and makes the probability of linear separability (and easy learnability) greater than 0.

Such increased representational ability is an advantage with working in Binary World, rather than using continuous vector components.

Another advantage of using a non-sparse binding matrix is that the representation decays more gracefully when noise is added to the matrix. Finally, in the nervous system, the majority of neurons synapse with many other neurons rather than a single neuron, making a permutation matrix appear much less neurally plausible.

- An important performance tuning issue for practical implementations is scaling the binding operator so that, for example, an $(\mathbf{M})^2\ \mathbf{V}^{\mathbf{girl}}$ term does not dominate other terms. One approach is to normalize the result of a binding operation so that the resulting vector has the same length as a vector for a single term, $\sqrt{\mathbf{D}}$. Alternatively, the normalization can make each $\mathbf{M}\ \mathbf{V}^{\mathbf{i}}$ phrase component have length $\sqrt{\mathbf{D}}$. Finally, we could just work in "Binary World," in which case the problem goes away.

## 3e. Multiple Simultaneous Representations

An important technique for reducing "brittleness" of the structure representation (such as parse information) is to simultaneously encode *several* structure descriptions (with different binding operators) in the vector by adding them. This increases robustness by having different structures "voting" in the final representation.

An example of multiple simultaneous representations is representing sentences as structureless additions of word vectors, plus binding of phrases, plus sequentially binding phrase components to fix their precise order. For example, with "*The smart Brazilian girl …*", we might have

$$\mathbf{(the + smart + Brazilian + girl) +}$$

$$\mathbf{M^{Actor}\ (the + smart + Brazilian + girl) +}$$

$$\mathbf{M^{Actor_Ordered}\ (the + M(smart + M(Brazilian + M(girl)))).}$$

We may also specify different positive weights for each of the three groups, for example to increase the importance of the top "surface" group with no binding.

Multiple simultaneous representations are helpful because we cannot know, a priori, which kind of phrase grouping will be critical for capturing the essence of what is to be learned in later stages.

For another example, if parser A results in sentence representation $\mathbf{V^A}$, and parser B produces $\mathbf{V^B}$, then the final representation for the sentence can be $\mathbf{V^A + V^B}$.

As a third example, if we have two (or more) image feature extraction programs (perhaps operating at different scales in the image), each program's outputs can be converted to a vector and then added together to get the final vector representation.

To summarize this Section, the two operators + and #, coupled with representing complex structure by applying # to additive phrases, permit us to represent objects, structures and sequences in MBAT. In Section 6, we check whether MBAT satisfies the representational constraints we have posed.

# 4. Learning and Representation: Three Stages

For both computational and neural systems, we distinguish three Computational Stages: *Pre-Processing, Representation Generation*, and *Output Computation*. This distinction is helpful, because learning plays a different role in each Stage.

## Pre-Processing Stage

The Pre-Processing Stage occurs prior to actually generating a vector representation. Here is where vector representations for objects (words, images) are developed so that similar objects have similar vectors. Typically the mapping is the result of a preliminary learning phase to capture object similarities in vectors (as discussed in Section 6).

As an important example of a Pre-Processing Stage in living neural systems, there appears to be much feed-forward processing of features of various complexity (eg., line detectors, moving edge recognizers, etc). These computations can be genetically hard-wired and/or learned during development, but then do not need to be re-learned in the course of the following Representation Generation Stage.

For automated systems, the identification of phrases (or "chunks") is a typical pre-processing operation that can be quite helpful for following Stages.

Although the learning involved in the Pre-Processing Stage may be computationally intensive, it is done only once, and then can be used in an unlimited number of representation calculations. Thus it avoids violating the Efficient Encoding Constraint, because it is not a part of the Representation Generation Stage. The features resulting from this Stage serve as the *inputs* to the representational system.

## Representation Generation Stage

Although the Pre-Processing Stage (and Output Computation Stage) can involve significant machine learning, there are reasons for the Representation Generating Stage to *avoid* machine learning.

- Internal learning means tailoring the representation for one set of applications, but this can make the representation less suited to a different set of problems.

- When representing inputs, a learning step might slow down processing so much as to make the resulting representation system impractical, thereby violating the Efficient Encoding Constraint.

On the other hand, we can envision a good case for learning some vector representations as part of a separate "long term memory" component, where we want to incrementally add a set of sequences to an existing set so that they may be recalled. Memory, recall, and novelty detection are important issues, but beyond the scope of this Representation paper.

## Output Computation Stage

Finally the Output Computation Stage is clearly a place where learning is vital for mapping representations to desired outputs. Here is where we benefit from being able to use conventional fixed-length vectors as inputs.

One major benefit is that a lot of previous theory is immediately applicable. These include the Perceptron Convergence Theorem [Rosenblatt 1959, see also Minsky & Papert 1969], Perceptron Cycling Theorem [Minsky & Papert 1969, Block & Levin 1970], Cover's theorem for the likelihood of a set of vectors to be separable [Cover 1965], and Vapnik-Chervonenkis generalization bounds [1971]. This body of theory permits us to prove learnability in many cases, as well as to set bounds on generalization.

To take a specific example, suppose we use random vector representations of words, and we have a collection of sentences with at most four phrases encoded as in Figure 2, and each sentence either contains the word "*girl*" or the word "*elephant*" (but not both). Then we can *prove* that perceptron learning will learn to distinguish the two cases, making a bounded number of errors in the process.

**<u>Proof</u>:** Consider

$$[\,(\,(\mathbf{M})^0 + (\mathbf{M})^1 + (\mathbf{M})^2 + (\mathbf{M})^3)\,(\,\mathbf{V}^{\mathbf{girl}} - \mathbf{V}^{\mathbf{elephant}}\,)\,] \bullet \mathbf{V}.$$

Excluding noise with mean 0, if **V** has $\mathbf{V}^{\mathbf{girl}}$ in a phrase, the dot product will be positive. If, by contrast, **V** has $\mathbf{V}^{\mathbf{elephant}}$ in a phrase, then the computation will be negative. Therefore the vector in brackets is a perfect linear discriminant. Now we can apply the Perceptron Convergence Theorem [Rosenblatt 1959, see also Minsky & Papert 1969, Gallant 1993] to know that Perceptron Learning will find some error-free classifier while making a bounded number of wrong guesses. (A bound is derivable from the bracketed term.)

This proof illustrates a simple and general way of proving that these kinds of mappings are learnable using the MBAT representational framework. We merely show that at least one error-free linear classifier exists, and then we can immediately conclude that perceptron learning will learn some error-free classifier (perhaps a different one) in finite time.

Note that there is no need to decode (i.e., fully recover) vectors in the process of learning!

Reviewing the three Stages, the overall processing picture is:

- a (one time, done in the past) *Pre-Processing Stage*, which may involve significant machine learning, feature computations, novelty detection, etc., which (efficiently) produces the inputs to:

- the *Representation Generating Stage* (our main focus), where there may be no learning, and followed by:

- an *Output Computation Stage*, which almost always involves machine learning for a particular task.

# 5. Capacity: Analytics and Simulation

For practical systems, we need to know what dimension, **D**, is required for vectors to represent objects and relations, and how **D** scales with increased system sizes.

The Appendix derives analytic estimates when adding **S** random +1/-1 vectors (referred to as the *bundled* vectors) to form vector **V**, and where **N** other random vectors are present in the system. (A bundle can be the vectors in a phrase.) In particular, we derive bounds and estimates for the required dimension, **D**, so that at least 98% of the time each of the **S** bundled vectors has a higher dot product with **V** than each of the **N** other vectors. Said differently, we seek error-free separation performance at least 98% of the time.

For a "Small" system where **S** = 20 bundled vectors, and where there are 1,000 other random vectors, we derive that **D** = 899 dimensions *guarantees* error-free performance at least 98.4% of the time. An example of a "Small" system application would be finding whether a simple diagram (collection of shapes in various configurations) is among 20 designated examples.

Similarly, for a "Medium" system with **S** = 100 bundled vectors, and where there are 100,000 other random vectors, we derive an estimate for the required dimension **D** of 6,927 for error-free performance 98.2% of the time.

Finally, for a "Large" system with **S** = 1,000 bundled vectors, and where there are 1,000,000 other random vectors, we derive an estimate for the required dimension **D** of 90,000 for error-free performance 99% of the time.

The approximation derived in the Appendix allow us to say how required dimension **D** scales as **S** and **N** increase. In summary, for a given error threshold:

- For fixed number of vectors **S** bundled together, as dimension **D** increases, the number of other vectors, **N**, we can distinguish from the bundled vectors (while keeping the same error threshold) increases *exponentially* with **D**.

- For fixed number of other vectors, **N**, as dimension **D** increases, the number of vectors **S** we can bundle together while distinguishing bundled and random vectors (and while keeping the error threshold) increases *linearly* with **D**.

Thus representing additional *bundled* vectors (**S**) is fairly "expensive" (required **D** is linear in **S**), while distinguishing the *bundled* vectors from **N** other *Random* vectors is fairly "cheap" (required **D** is logarithmic in **N**).

In addition to the analytic estimates in the Appendix, we also performed capacity simulations for "Small" and "Medium" sized problems. Here we investigated, for different values of vector dimension **D**, computing the sum of **S** vectors to form bundled vector **V**. We then found the **S** vectors with highest dot products with **V** among the **S** bundled vectors and the **N** additional random vectors. We computed:

1. The fraction of *bundled* vectors that are in the **S** vectors having highest dot product with **V**.
2. The fraction of *trials* that produce error-free performance: all *bundled* vectors have higher dot products with **V** than any of the *Random* vectors. (This is the same measure analyzed in the Appendix.)

Figures 3 and 4 give the results. The “Small Sized” system required several hours computation time on a 2.4 GHz laptop, and the “Medium Sized” system required 24 hours simulation time.

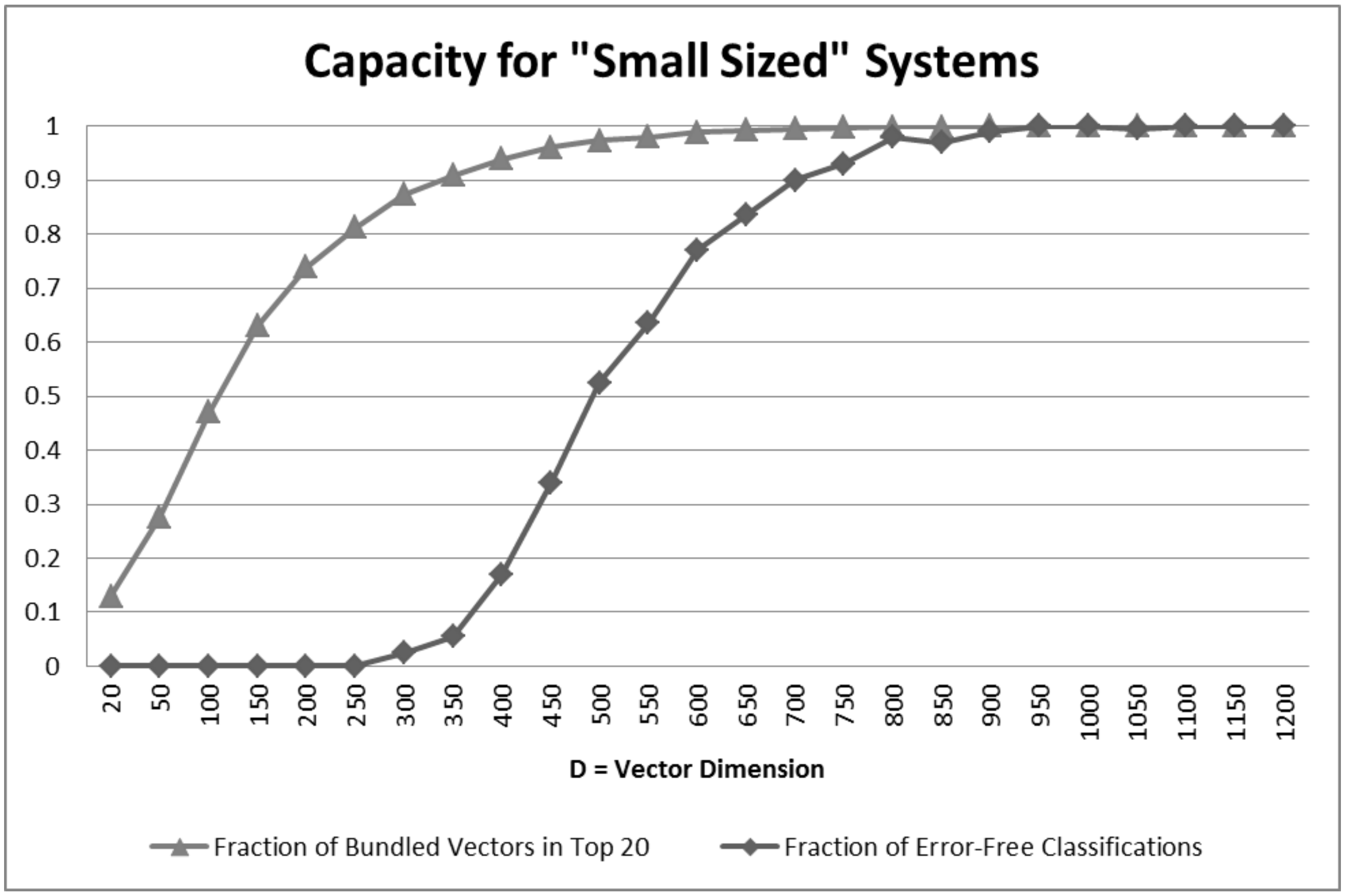

**Figure 3: Capacity simulation for a "Small" system consisting of S=20 vectors *Bundled* together to form vector V, plus N=1,000 additional *Random* vectors.** When computing the top 20 vector dot products with V, the top series shows the fraction of *bundled* vectors in the top 20, and the bottom series shows the fraction of error-free separations (all *bundled* vectors have higher dot products with V than all *Random* vectors). Averages over 200 trials.

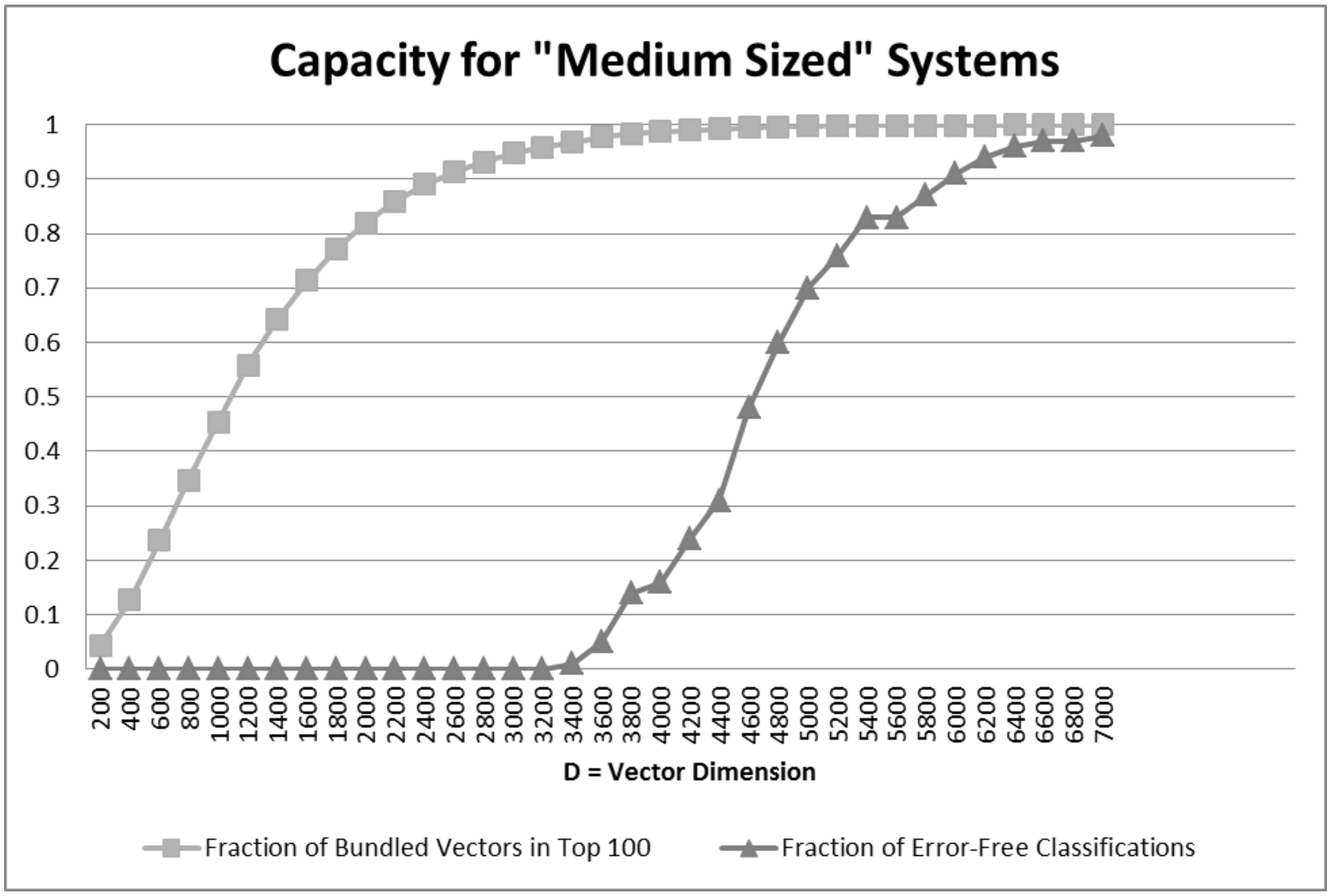

**Figure 4: Capacity simulation for a "Medium" system consisting of S=100 vectors *Bundled* together to form vector V, plus N=100,000 additional *Random* vectors.** When computing the top 100 vector dot products with V, the top series shows the fraction of *bundled* vectors in the top 100, and the bottom series shows the fraction of error-free separations (all *bundled* vectors have higher dot products with V than all *Random* vectors). Averages over 100 trials.

From the simulations, we conclude:

- For a "Small" system (**S**=20; **N**=1,000), a *lower* estimate for required vector dimension **D** is only 350. This gives 90% (18 of 20) of the top vectors being *bundled* vectors.
- For the same "Small" system, an *upper* estimate for required vector dimension **D** is 900. This gives 98% probability of having *error-free performance* with highest dot products all being the 20 *bundled* vectors.
- Similarly, for a "Medium" system (**S**=100; **N**=100,000), we have lower and upper estimates for required dimension **D** of 2,500 and 7,000 respectively.

In the Appendix, we derive an approximation formula for *p*, the probability of error-free performance. Letting

$\mathcal{T}$**(x)** = one-sided tail probability in a normal distribution of a random variable being at least **x** standard deviations from the mean

We derive

$$p = \mathbf{1 - NS}\ \mathcal{T}(\sqrt{(\mathbf{D}/(\mathbf{2S-1}))})$$

where the approximation is valid when *p* is close to 1 (ie., *p* > 0.9).

The analytic estimates for required dimensionality for error-free performance in the Appendix are in close agreement with simulation results. Results are summarized in Table 2.

| | | | From Simulations | | From Formula |
|---|---|---|---|---|---|
| System Size | S = Number of Vectors Bundled | N = Other Random Vectors | Required D for 90% of Closest Vectors Being in the Bundle | Required D for Probability >= .98 of Error-Free Performance | Required D for Probability >= .98 of Error-Free Performance |
| Small | 20 | 1,000 | 350 | 900 | 899 |
| Medium | 100 | 100,000 | 2,500 | 7,000 | 6,927 |
| Large | 1,000 | 1,000,000 | | | 90,000 |

**Table 2**: Simulated and estimated vector dimension required for (a) 90% of the **S** closest vectors being in the Bundle, and (b) at least .98 probability of error-free performance.

Some additional comments on capacity.

- It is possible to improve recognition ability for bundle vectors when the vectors added together are not random. For example, in natural-language applications we can use a "language model" [Brown et al. 1990], which gives statistics for a vector being included in the presence of other vectors in the bundle.

- When we use vectors such that similar objects have similar vectors, for example giving similar vectors to similar words, then this will decrease raw discrimination ability to the extent that vectors are more similar than random. However, this should help, rather than hurt, an implementation – that's the reason we made the objects more similar in the first place!

- When machine learning follows representation in a practical system, this may require significantly *less* dimensionality than required for error-free discrimination, depending upon the specifics of the learning task and performance requirements.

  For example, if there are **S** positive examples and **N** negative examples to be learned, we don't need **A** • **V** > **R** • **V** for every case where **A** is a positive example and **R** is a negative example.

  Instead of using the sum of the positive examples, **V**, to discriminate, we have the liberty of finding *any vector* **X** that does a good job of making **A** • **X** > **R** • **X** for each pair **A, R**. If any **X** exists that works for all pairs, it is a linear discriminant which is easily learned by Perceptron Learning. Moreover, most practical modeling problems do not require *perfect* discriminants.

Finally, it is worth estimating computational requirements. Generating a representation vector of dimension **D** involves a **D** × **D** matrix multiply for each binding operation, plus vector additions as needed, regardless of the number of objects in the system. For a "Medium" sized system with **D** = 2,000 to 7,000, generating a representation on a single processor computer is clearly practical, although each binding requires 8 billion to 343 billion multiplications and additions.

(Speedups are available using multiple processors and also by fast matrix multiplication techniques.)

For a "Large" system with **D** = 90,000, extensive parallelism would be required. In general, multiple processors divide the computation time by the number of processors for these computations. It is noteworthy that a living system with neurons summing synaptic inputs from many other neurons, and with neurons working in parallel, could conceivably compute binding operations of vector sums for the state update Equation 3.1 in a single time step!

Follow-on learning in the Output Computation Stage may require less computation. For example, perceptron learning requires only vector dot products and vector additions for each iteration, avoiding the more expensive matrix multiplications. Of course, the number of iterations required is also important. (Worst-case bounds on required iterations for linearly separable problems grow linearly with the dimension and the number of training examples, and grow by the square of the length of the shortest integral solution vector [Gallant 1993].)

We conclude that vector dimension requirements and computations appear easily manageable for practical implementations.

## 6. Checking the Constraints

We want to verify that by using distributed vectors and two operations on vectors, addition (+) and binding (#), plus MBAT's approach for representing complex structure, we are able to satisfy the Constraints from Section 2.

Constraints 1 (fixed-length vector) and 2 (distributed representation) are obviously satisfied. We have seen how the binding operator can represent structure (Constraint 3), including a practical solution to the binding problem, as well as sequences (Constraint 5). Computations are clearly linear in the number of objects, plus complexity (description size) of structures (Constraint 6).

Let us consider Constraint 4 (similar objects/structures map to similar representations). Although "similarity" is not precisely defined, nevertheless we can verify our benchmark test cases.

With natural language, there are a number of ways to get similar vector representations for similar words, as surveyed in Section 7.

Note that this (learned) preprocessing is a one-time computation that does not affect speed during the Representation Generation Stage.

Now, suppose we encode "*The smart girl saw the gray elephant*" by the sum of vectors:

$$\mathbf{V} = \mathbf{M}^{\mathbf{Actor}}\,(\mathbf{the} + \mathbf{smart} + \mathbf{girl} + \underline{\mathbf{phraseHas3words}}) + \mathbf{M}^{\mathbf{verb}}\,(\mathbf{saw} + \underline{\mathbf{phraseHas1word}}) + \mathbf{M}^{\mathbf{Object}}\,(\mathbf{the} + \mathbf{gray} + \mathbf{elephant} + \underline{\mathbf{phraseHas3words}}).$$

Then if we have object encodings such that similar objects have similar vectors, we can interchange similar objects and leave the resulting representation vector similar to the original. For example, if $\mathbf{V}^{\mathbf{intelligent}}$ is similar to $\mathbf{V}^{\mathbf{smart}}$, then "*The intelligent girl saw the gray elephant*" will have a vector very similar to **V**.

Turning to structure similarity, we need add only one additional vector term, $\mathbf{V}^{\mathbf{Passive_Voice}}$, to **V** in order to encode

“*The gray elephant <u>was seen</u> by the smart girl*”.

Thus these two representation vectors are very similar as vectors.

In a like manner, we can encode

“*The smart girl <u>I just met</u> saw the <u>young</u> gray elephant <u>eating peanuts</u>*”

by adding to the original vector those additional vectors for the two new clauses, and adding **young** to the elephant clause. Once again, we arrive at a vector similar to the original one (i.e., significantly closer than two randomly chosen sentence vectors).

The last remaining Constraint, Neural Plausibility, is even less precise. However, we maintain this Constraint is satisfied by having a system with individual vector components (neurons), a notion of state consisting of which neurons are firing, and a way to represent objects and structural relationships in the overall state that does not make unreasonable demands on “wiring” of the system by genetics or during growth.

# 7. Prior Research

Vector Symbolic Architectures were summarized in Section 1, and are further discussed in the next Section.

A key early source for Distributed Representations is [Hinton’s 1984], as well as [Hinton 1986]. This paper presents characteristics, advantages and neural plausibility arguments. (These topics are reviewed in Gallant [1993].)

In Information Retrieval, the use of high dimensional vectors to represent terms (words) was pioneered by Salton & McGill [1983]. Deerwester et al. [1990] represented terms, documents and queries by starting with a document-by-term matrix, and then using Singular Value Decomposition to reduce dimensionality. This also achieves some measure of similarity of representation among similar terms. A later approach used random vectors and learned modifications of those vectors to represent similarity of meaning among words in the MatchPlus system [Caid et al., 1995]. The basic idea was to start with random vectors for terms, and make passes over a corpus while modifying vectors by adding in a fraction of surrounding vectors (and normalizing).

In the same spirit, other language systems make good use of a “language model” to statistically compute probabilities of a word given its immediate predecessors [Brown et al. 1990], or computing the probability of a word given its surrounding window [Okanohara & Tsujii 2007, Collobert &Weston 2008]. See also the Brown clustering algorithm [1992], phrase clustering [Lin & Wu, 2009] and [Huang & Yates 2009]. Self-Organizing maps present another possibility [Kohonen 1995, Hagenbuchner et al. 2009].

However, all this Information Retrieval work is still within the bag of words limitations imposed by not having a binding operation.

More recently, Jones and Mewhort [2007] looked at incorporating positional information with semantic vectors created by a similar approach to MatchPlus. They capture order information by looking at surrounding windows (up to distance 7) for each term in a corpus. They then take the (HRR) convolution of each window, while replacing the target term by a dummy.

For example, suppose we're learning the order vector for King in "*Rev. Martin Luther King, Jr. said ...*". Then letting ϕ be a constant placeholder for the target term "*King*", we would add HRRs for: "*Luther* * ϕ", "ϕ * *Jr*", "*Luther* * ϕ * *Jr*", etc. The resulting order vector is then normalized and added to the semantic vector for King. The result is a vector that captures semantics as well as word order syntactic effects. Similar results were obtained by Sahlgren et al. [2008] by encoding order information with permutations; see also Recchia et al. [2010]. Such vectors should provide interesting starting codings for terms in language systems, including MBAT.

With respect to matrix multiplication bindings, Hinton's "triple memory" system [1981] used random matrix connections in a subsidiary role while focusing on learning, rather than representation. Also, Plate's book [2003, page 22] later mentions in passing exactly the "Two Input" version of the binding operator from Section 3, which he attributes to Hinton. Plate also lists matrix multiplication as an alternative binding possibility in Section 7.3, Table 26.

In a Computational Linguistics setting, Rudolph & Giesbrecht [2010] proposed using only matrices (rather than vectors) to represent objects, and examined matrix multiplication as a composition operation. Similar results were obtained by Sahlgren et al. [2008] by encoding order information with permutations; see also Recchia et al. [2010]. However, vector addition carried out by sparse matrices in $\mathbf{D}^2$ dimensions rather than $\mathbf{D}$ dimensions is inefficient. There is also loss of the binding recognition property once we use a large number of different matrices for multiplication, rather than a small set of matrices for binding operators.

Mitchell & Lapata [2008], also in a Linguistics domain, mention the binary version of the # operator in passing, although most of their efforts focus on a bag-of-words semantic space model.

Turning to sequences, the traditional approach to dealing with sequential inputs (eg., a sentence) is to use a sliding window. A related approach, Elman nets [1990], are three-layer neural networks that copy the hidden layer outputs as net inputs for the next cycle, thereby producing an additional "sliding window over hidden layer outputs." Elman nets are therefore able to accumulate some state information over the entire sequence.

Another related sequence approach, Time Delay Neural Networks of Waibel et al. [1989], has several layers of groups of hidden nodes. The first node in each group sees (say) nodes 1-3 in the group (or input) immediately below, the second node of each group sees nodes 2-4 in the group below, etc. Thus we have a multi-stage fan-in from the input layer, with each stage reducing dimensionality while expanding global coverage.

All three of these approach typically employ Backpropagation (or variants) to adjust weights in response to training data. Therefore they are more approaches for learning algorithms, rather than approaches for representation. Although we could consider hidden layer activations as representations for new inputs after learning has ended, there is limited ability to recognize stored objects, and the only type of structure that is explicitly captured is sequentiality. Nevertheless, these techniques might prove useful in a Pre-Processing Stage prior to generating representations.

For sequence representations that do not require learning, Kanerva [1988] represents sequences using pointer chains. Later, Plate [2003] employs trajectory association, where the idea is to bind powers of a vector to sequence information. For example if we want to represent the sequence A, B, C, we can take some fixed vector **V** and compute

$$\mathbf{V*A + V*V*B + V*V*V*C}.$$

There are additional variations involving helper terms for easier decoding.

There is also a body of research on learning with structural inputs, much of which involves using Backpropagation related algorithms to learn weights in a pre-defined network without directed loops [Frasconi et al. 1998]. Again, the focus is on learning, rather than representation. The Backpropagation computations (along with potentially large numbers of hidden units) make this approach impractical for generating general-purpose representations.

Another early work involving Reduced Representations is Pollack's RAAM (Recursive Auto Associative Memory) architecture [1990] and later extensions, for example LRAAM (Labeling RAAM) by Sperduti, Starita & Goller [1995]. These approaches use Backpropagation learning (or variants) on a network of inputs, hidden units, and outputs that attempt to reproduce inputs. The hidden units, after learning, encode the reduced representations of the inputs. A drawback of these approaches is the need for learning over all inputs to achieve the representations of the inputs. For example, adding additional input cases requires re-learning the representation for all previous input patterns using Backpropagation (violating the Efficient Coding Constraint). Improvements in capacity and generalization were reported by Voegtlin & Dominey [2005]. Although these approaches are all too slow (non-linear) for the Representation Generation Stage, their abilities to capture generalization may present good synergy as part of the Pre-processing Stage.

Another important line of research for learning structures with generalization was Hinton's family tree tasks [1986, 1990], followed by Linear Relational Embedding [Paccanaro & Hinton 2001a, 2001b; Paccanaro 2003]. As with RAAM architectures, generalization ability may prove useful in producing Pre-processing Stage inputs for MBAT or other approaches.

Sperduti [1997] proposed a "generalized recursive neuron" architecture for classification of structures. This complex neuron structure can be seen as generalizing some other ways of encoding structure, including LRAAM, but representation size grows with respect to the size of the structure being represented, as does computational requirements.

Another recent approach represents structures by dynamic sequences, but requires a Principal Component Analysis (PCA) for obtaining the representation. Sperduti [2007] reports a speed-up for PCA calculation; the result could play a role in Pre-processing Stage learning of inputs, but is still too computationally demanding for computing representations in the Representation Generation Stage.

It is worth noting that Sperduti et al. [1995] conduct *simulations* to show good performance for learning to discriminate presence of particular terms in the representation. By contrast, Section 4 *proves* learnability for a similar task, without the need for simulation.

More recently, Collobert & Weston [2008] show how a general neural network architecture can be simultaneously trained on multiple tasks (part-of-speech Tags, chunks, named entity Tags, semantic roles, semantically similar words) using Backpropagation. They encode sentences using

Time Delay Neural Networks of Waibel et al. The network learns its own vector representations for words during multi-task learning, so that different tasks in effect help each other for a shared portion of the vector representation for each task. However, each task ends up with its own vector representation for the non-shared part. Outputs consist of either a choice from a finite set (eg., part-of-speech Tags) or a single number (probability) on a word-by-word basis.

It is not apparent that their internal vector encodings can serve as representations for, say, sentences, because each learning task produces a different vector for the same sentence. Nor are the sets of outputs, produced for each word, easy to directly convert into a fixed-length vector for the sentence.

However, there is an appealing natural synergy of Collobert & Weston's system with the representation we examine, because outputs from their system can serve as structure information to be included in the representation vector. In particular, the "chunking" outputs can comprise phrases, which are ideal candidates for binding operations in constructing the vector for a sentence. Vector Symbolic Architectures in general, and MBAT in particular, give a natural way to leverage word-by-word information as inputs to learning algorithms by converting it to fixed-length, distributed vectors.

In a later work, Collobert et al. [2011] develop a unified neural network architecture for these linguistic tasks, where all tasks are trained using two somewhat complex architectures based upon Time Delay Neural Networks. As in previous work, their system outputs a set of tags for each word. Training time is one hour to three days, depending upon the task, and scoring of new input is very fast after training. Their impressive point is that, at least for producing various Tags for words in the domain of language, one of two neural network learning architectures can produce excellent results for a variety of tagging tasks. The architecture is highly tuned to producing word-by-word Tags from language input. Therefore, it seems very hard to adapt this approach to other tasks such as machine translation, where more general output structures must be produced, or to other domains such as image processing, without first converting outputs to fixed-length vectors.

These papers are example of a broader category, Structured Classification, where for each sequential input object we compute either one choice from a fixed and finite set of choices, or we compute a single scalar value. There is much other research in the Structured Classification paradigm, which we do not review here.

Recently Socher et al. [2010, 2011a, 2011b] have shown how binding matrices can be *learned* using Backpropagation with complex loss functions. Socher's Recursive Neural Networks (RNN) are binary trees with distributed representations, which are structurally identical to "Two Input" binding operators in Section 3. In particular, Socher's matrix [2011a] for combining two vectors is equivalent to concatenating rows from $\mathbf{M}^{Left}$ and $\mathbf{M}^{Right}$ to form a single "double wide" matrix for applying to pairs of concatenated column vectors.

The RNN approach is applied to Penn Treebank[3] data to learn parsing and the Stanford background dataset[4] to obtain a new performance level for segmentation and annotation. They also report [2010] excellent results with the WSJ development dataset, and an unlabeled corpus of the English Wikipedia [2011b].

[3] http://www.cis.upenn.edu/~treebank/
[4] http://dags.stanford.edu/projects/scenedataset.html

Socher's work demonstrates that the kind of Vector Symbolic Architectures we describe can be useful for practical problems.

Finally, there is interesting work by Maass et al. [2002] on asynchronous modeling of noisy neuron behavior, including recurrent connections, in their "Liquid State Machine" model. They show ability to train output neurons to discriminate noisy input patterns in neurons with fully asynchronous firing times.

This modeling is at a granular neuron level, and is therefore more suited for neural modeling and less suited for large scale, practical systems. For example, simulations typically deal with distinguishing among a small number of input patterns, and there is no attempt at explicitly representing complex structure among objects.

Nevertheless, the "Liquid State Machine" models of Maass and colleagues share several characteristics with the kind of representational systems we examine, including:

- They are general representations, not tuned to any specific task.
- There is a specific task/output readout Stage that involves learning.
- State transitions are similar to Equation 3.1 in Section 3.
- The overall system does not need to converge to a stable state, as with most learning algorithms.
- The mathematical model can be used to hypothesize computational explanations for aspects of neural organization and processing.

Maass et al. investigate "local" versus "long range" recurrent connections, giving computational explanations for the distributions. They find that less than complete connections work better than complete connections with their asynchronous, noisy, dynamic models. (This result may not apply to non-asynchronous systems.)

# 8. Discussion

We have shown that a desire to apply standard machine learning techniques (neural networks, perceptron learning, regression) to collections of objects, structures and sequences imposes a number of Constraints on the *Representation* to be used. Constraints include the necessity for using distributed, fixed-length vector representations, mappings of similar objects and structures into similar vector representation, and efficient generation of the representation.

In response we have developed MBAT, a neurally plausible Vector Symbolic Architecture that satisfies these constraints. MBAT uses vector Addition and several possible variants of a vector Binding operator, plus a Complex Structure Methodology that focuses upon additive terms (ie., phrases).

## MBAT as a Vector Symbolic Architecture

Viewed from the perspective of Vector Symbolic Architectures, MBAT can be characterized as follows:

- Vector components are either continuous, or are two-valued (eg., +1/-1).

- Addition is vector addition, optionally followed by thresholding as in Binary Spatter Codes.

- Binding is a unary operator consisting of matrix multiplication. Either one matrix or a matrix chosen from a small set of matrices ($\mathbf{M}^{\mathbf{Actor}}$, $\mathbf{M}^{\mathbf{Object}}$, etc) is used for binding. Components of matrices can be chosen at random, and can have +1/-1 entries. If vectors are restricted to having +1/-1 components, then matrix addition and multiplication are followed by a thresholding step. Another variation adds binding operands back into the result of matrix multiplication.

  A two-argument version of binding is available by multiplying each argument by one of two fixed matrices, and adding the results.

- Quoting is by repeated matrix multiplication, or by using different matrices from a small set of matrices.

  This is similar to quoting by permutation, for example Gayler [2003], Kanerva [2009], and Plate [2003], except we need not restrict ourselves to permutation matrices. (See comments in Section 3 on permutations.)

- The Complex Structure Methodology applies (unary) binding to *additive phrases*, **M(V+W+X)**. Each of the added terms may be the result of another binding operation. Several different representations can be simultaneously represented by adding their vectors.

## Vector Symbolic Architectures and Complex Structure Methodology

The procedure for encoding complex structure deserves further comment with respect to VSAs in general, and Holographic Reduced Representations in particular.

For VSAs, Context Dependent Thinning will map similar structures to similar vectors, as required by Constraint 4. Other VSA methods can run into problems with this constraint: there is no vector similarity between **V*W** and **V*W*X**. For example, **(smart * girl)**, and **(smart * Brazilian * girl)** have no similarity.

Another troublesome case for all VSAs is representing repeated objects whenever binding is used for phrases. We might like **(tall * boy)**, **(very * tall * boy)**, and **(very * very * tall * boy)** to each be different, yet with appropriate similarity between pairs of phrases. However HRRs give no relation between pairs of phrases, and BSC gives no difference between phrases 1 and 3 (unless various workarounds are employed).

The issue is whether to represent phrases by binding the terms directly, **(V*W*X)**, or by binding their sum, **#(V+W+X)**, as in MBAT.

Note that in HRRs we can convert a two-argument ***** operator to a unary ***** operator by

$$\mathbf{*(V+W+X) \equiv dummy * (V+W+X)}.$$

(There are other ways to create unary operators from two-argument operators.)

When included in representations, such additive phrases permit HRRs to map similar phrases to similar vectors, thereby satisfying Constraint 4 in Section 2. We now have

***(smart + Brazilian + girl)**

similar to

***(Brazilian + girl)**,

which was not the case with **(smart*Brazilian*girl)** and **(Brazilian*girl)**. Computation is also reduced, because ***** requires more work than computing vector addition.

Moreover, it is also much easier to recognize whether $\mathbf{V}^{\mathbf{girl}}$ is in ***(smart + Brazilian + girl)** than it is to recognize whether it is in **(smart * Brazilian * girl)**, because we can take a single dot product, similar to equations 3.2 – 3.4.

Thus employing an additive phrase as the argument for a unary binding operator would appear beneficial for HRR representations (and also possibly Socher's models). Even better, we can *combine* an additive phrase vector with an HRR binding (rather than *replacing* it) as in

**(V*W*X) + dummy*(V+W+X)**.

This is an example of multiple simultaneous representations.

Finally, it can be seen that HRR binding of sums, as in **dummy*(V+W+X)**, corresponds precisely to MBAT bindings, where the MBAT binding matrix is restricted to matrices having each row other than the first be a single rotation of the preceding row.

## Jackendoff's Challenges

These complex structure issues connect to Jackendoff's Challenges to Cognitive Neuroscience [2002] and to Gayler's response [2003]. Jackendoff issued four challenges for Cognitive Neuroscience, of which Challenge Two, "the problem of 2", involves multiple instances of the same token in a sentence, for example "*the little star*" and "*the big star*". We want to keep the two stars distinct, while maintaining a partial similarity (both are stars). $\mathbf{M}^{\mathbf{Actor}}$ **(the + little + star)** and $\mathbf{M}^{\mathbf{Object}}$ **(the + big + star)** are different, yet both are similar to ($\mathbf{M}^{\mathbf{Actor}} + \mathbf{M}^{\mathbf{Object}}$) **star**, as desired.

Further, using additive phrases, we can encode the original example discussed by Jackendoff, "*The little star's beside a big star*" by the vector

$\mathbf{M}^{\mathbf{Actor}}$ **(the + little + star)** + $\mathbf{M}^{\mathbf{Verb}}$ **( 's )** + $\mathbf{M}^{\mathbf{Relation}}$ **(beside + the + big + star)** and similarly for HRRs using additive terms.

## Applications

For applications of MBAT, each modeling environment will require that we select the appropriate Preprocessing Stage details to take advantage of specific characteristics. For example with language, we need to select a method for making similar terms have similar vectors, decide which chunking, tagger (etc.) software is available for specifying structure information, and decide which binding operator variation to employ.

Similarly, for machine vision, we need to see what are the available feature detectors and what higher-level structure information is available.

However once these details are specified, we can create representations using MBAT, and then use standard machine learning algorithms directly "out of the box" to learn tasks of interest.

We believe that MBAT provides a practical playing field where machine learning can efficiently operate upon objects, their structures, and sequences all at once – as either inputs or outputs.

Let us briefly look at possible applications.

- For Information Retrieval, representing structure (in addition to terms) may improve performance in various learning tasks, for example finding specific categories of articles (eg., "*joint ventures where a specific agreement is announced*").

- In natural-language processing, MBAT gives a way to make good use of parse information keyed to sets of *phrases* as in

  $$\mathbf{V} = \mathbf{M}^{\mathbf{Actor}}\,(\mathbf{the + smart + girl} + \underline{\mathbf{phraseHas3words}}) + \mathbf{M}^{\mathbf{verb}}\,(\mathbf{saw} + \underline{\mathbf{phraseHas1word}}) + \mathbf{M}^{\mathbf{Object}}\,(\mathbf{the + gray + elephant} + \underline{\mathbf{phraseHas3words}}).$$

  Thus we have a direct approach, using existing Taggers, for learning machine translation from paired corpora with paired sentences. For example, we can work from a collection of English sentences with corresponding French translations. (Different dimension vectors can be used for the two languages.) We take the vectors for English and French translations, and then train a classifier to go from the **D** components of the French sentence vector to the first component of the English vector. Similarly for the other components of the English vector, resulting in **D** classifiers in total. The net result is a map from French words and parse structure to English words and parse structure. Whether this would work well, or whether it would not work at all, would need to be explored in an implementation.

  A potential difficulty with translation is that it may be challenging to construct an output module that goes from a vector to a corresponding string of terms. For this task, we need to *recover* the sentence, rather than *recognize* the components of a sentence encoded by a vector. Here it is likely that embedding Tags into phrases (eg., **phraseHas3words**) will help with output. Nevertheless, constructing a vector-to-sentence module is a critical – and likely difficult – task for translation (or summarization) when using vector representations.

  There are other potential applications that share similarities with representing natural language, including representing genome sequences and chemical structures.

- Another application area is computer vision. Here a natural problem seems to be recognizing whether an image contains a specific object. Such tasks likely require multiple representations of structures at different scales in the image. This suggests combining multiple feature detectors (working on different scale sizes), and employing different binding operators ($\mathbf{M}^{\mathbf{close_to}}$, $\mathbf{M}^{\mathbf{above}}$, etc) to end up with sums of terms such as:

$\mathbf{M^{close_to}}$ **(location + shape_1 + shape_2)** and

$\mathbf{M^{above}}$ **(location + shape_1 + M shape_2),** etc.

As with natural language, our representation only gives a research path to generating a practical system, but does not totally solve the problem. A creative implementation is still required.

- A final application is neural modeling. In particular, we want to capture the computational essence of neural information processing at a useful level of mathematical abstraction.

  Of course the brain does not have complete recurrent connections, where every neuron is connected to every other. (In other words, binding matrices contain many zero terms, which doesn't fundamentally change the analysis.) Specialized brain sub-structures and many other details are also ignored in our abstract model.

  Nevertheless, the MBAT computational architecture suggests a number of computational explanations for large-scale aspects of neural organization.

  The most important example is that the binding operation suggests a plausible *computational* reason for the brain having so many *recurrent connections*. (A neuron has an estimated average of 7,000 synaptic connections to other neurons.)

  A second, and more subtle, computation-based explanation is for an aspect of neural organization currently taken completely for granted: the need for a *separate memory mechanism*. In other words, why not have a unified "whole brain" that simply remembers everything? The computational explanation is that, with the representation we have developed, objects/structures are expensive to store, because the number of required vector components rises linearly with the number of stored items. Also, *recovery* – as opposed to *recognition* – of objects is not directly given by the representation we have explored. Hence the need for a specialized memory functionality, separate from general binding, that efficiently stores large numbers of objects and that facilitates recovery of stored vectors.

  Finally, the MBAT architecture can motivate Cognitive Science hypotheses. For example, we can hypothesize that there are neural resources devoted to recognizing *phrases* in language at an early processing stage. This hypothesis is supported by the computational benefits we have seen, as well as by the help given for phrase recognition in both written and spoken language: punctuation, small sets of prepositions and conjunctions, typical word ordering for phrases, pauses in speaking, tone patterns, coordinated body language, etc. Recent Magnetoencephalography studies by Bemis & Pylkkänen [2011] give additional support.[5]

For some applications, VSAs in general will need to be augmented by several very important missing pieces. These revolve around learning (including long term and short term memory), storage/recall of vectors, novelty recognition and filtering, focus of attention, and dealing with large sequences or structured large objects (reading a book). Is there a good extension of MBAT that will properly accommodate these functions?

[5] Localized neural response for minimal phrases ("red boat") occurs not in traditional language areas, but instead first in areas associated with syntax, followed by areas associated with semantic processing.

We leave such Representation Theory development, as well as construction of practical systems in natural language, computer vision, and other areas, to future work.

## Acknowledgments

Thanks to Tony Plate, Pentti Kanerva, Ron Rivest, Charles Elkan, Bill Woods, and Bob Cassels for useful discussions and suggestions on an early draft, and to the Referees for extremely helpful comments.

# Appendix:  Capacity for Vector Sums

It is possible to derive good approximation formulas for storage and recognition performance within a single distributed vector, as we show below.

Previous approximations appear in the Appendices in Plate's book [2003] and Anderson [1973], both of which look at vector components drawn from normal distributions rather than +1/-1. Their findings for normally distributed components agree with Propositions 1-3 below.

We use the following notation, assumptions, and simplifications:

- **D** = dimension of vectors
- **S** = number of vectors bundled together to form vector **V**
- **N** = number of randomly generated vectors that we wish to distinguish from those used in the sum forming **V**
- $\mathcal{T}$**(x)** = one-sided tail probability in a normal distribution of a random variable being at least **x** standard deviations from the mean
- **Z** = $\sqrt{(D/(2S-1))}$ for fixed **D** and **S**
- All object vectors are randomly generated +1/-1 vectors.
- For simplicity, we do not include continuous vectors produced by binding operations (vectors formed by random matrix multiplications).  We could, however, include such vectors if we're thresholding matrix multiplication results to obtain +1/-1 vectors.

The first thing to note is that the dot product of two random vectors has mean 0 and variance **D** (and hence standard deviation √**D**), because it is the sum of **D** independent random variables, each with mean=0 and variance=1.

Similarly, when we add **S** vectors to form vector **V**, then for a random vector **R,** we have  **R • V** also has mean 0 and variance **SD**, giving standard deviation √**(SD**).

Let **A** be a randomly chosen vector from the **S** vectors <u>**A**</u>dded (bundled) to form **V**, and **R** a <u>**R**</u>andom vector.

We're interested in the probability of a mistaken recognition where **R • V > A • V.**

As in the proof of Property 1 for vector addition,

**R • V = 0 + <mean 0 noise from the dot product with a sum of S vectors>**, and

**A • V = D + <mean 0 noise from the dot product with a  sum of S-1 vectors>.**

For **D** large enough, the Central Limit Theorem of statistics guarantees that the first noise term will be closely approximated by a normal distribution with mean 0 and standard deviation **√(SD)**, denoted $\mathcal{N}$**(0, √(SD))**. Similarly, the second noise term is closely approximated by $\mathcal{N}$**(0, √((S-1)D))**

So the probability of a mistake with these two vectors is given by:

> the probability of a random value selected from $\mathcal{N}$**(0, √(SD))** being greater than a random value selected from $\mathcal{N}$**(D, √((S-1)D).**

This is equivalent to the probability of a random vector being negative when selected from $\mathcal{N}$**(D, √((2S-1)D)**, because the difference of two normal distributions, **X-Y**, is a normal distribution having mean equal to the difference of the two means, and variance equal to the sum of the two variances. [Many thanks to the Referee who pointed this out.]

> Proof: From basic definitions, if **Y** is normally distributed then so is **(-Y)**, with mean**(-Y) =** -mean(**Y**) and with Var(**-Y**) = Var(**Y**). The result now follows from well-known properties for the sum of two normal variables, applied to **X** and **(-Y)**.

Thus, looking at standard deviations, an error occurs when a difference is in the tail probability at least **D / √ ((2S-1)D) = √ (D/(2S-1))** standard deviations from the mean.

Here it is convenient to introduce some simplifying notation. We define:

> **Z** = √ **(D/(2S-1))** for fixed **D** and **S.**

Thus, for pre-specified **D** and **S**, we have **Z** corresponding to **D** as measured in standard deviations of the difference in noise terms.

We also adopt the notation $\mathcal{T}$**(x)** = one-sided tail probability of a random variable being at least **x** standard deviations from the mean in a normal distribution.

Thus an estimate for the error with the pair of randomly chosen vectors, one from the bundled vectors and some other random vector is

$$(*) \qquad \mathcal{T}(Z) = \mathcal{T}(\sqrt{D/(2S-1)}).$$

Now we only need to note that the tail probabilities of a normal distribution decrease exponentially (ie., as $e^{-x}$) with the number of standard deviations to conclude:

*Proposition 1:* For a fixed number, **S**, of random +1/-1 vectors bundled together to get vector **V**, the probability of a random vector having greater dot product with **V** than a randomly selected vector in the sum decreases exponentially with vector dimension, **D**.

*proof:* $\mathcal{T}$**(√ (D/(2S-1)))** decreases exponentially as **D** increases.

We are really interested in the probability of *all* **S** of the vectors in the sum having greater dot product with **V** than *any* of **N** random vectors. This probability is given by

$$[\, 1 - \mathcal{T}(Z)\, ]^{NS} \approx 1 - NS\, \mathcal{T}(Z)$$

assuming **NS** $\mathcal{T}$**(Z)** is close to 0, and dropping higher-order terms. Thus we have

<u>*Proposition 2*</u>*:* For a fixed number, **S**, of random +1/-1 vectors bundled together to get vector **V**, if we generate **N** additional random vectors, then the probability that each random vector has less dot product with **V** than each vector in the sum equals 1 minus an exponentially decreasing probability as vector dimension, **D** increases.

In other words, as **D** increases, the probability of less than error-free discrimination between **S** vectors in the bundle and **N** random vectors decreases exponentially.

<u>*Proposition 3*</u>*:* For a fixed number, **N**, of random +1/-1 vectors, the number of vectors, **S**, in a bundle that can be perfectly discriminated against with constant probability increases nearly linearly with vector dimension **D**. (More precisely, the required **D** increases linearly with **S**, plus second order terms**.**)

In summary, as **D** increases, the number of random vectors we can discriminate from a sum of **S** vectors increases exponentially, and the number of vectors we can have in the sum while maintaining discriminatory power increases slightly less than linearly with respect to **D**.

It is instructive to compute several of these bounds.

For a "Small" system where we sum **S**=20 vectors (ie., terms, Tags, image primitives, etc.) to form **V**, and we have **N** = 1,000 additional random vectors, we compute

$$\text{Probability of error} = (20)\,(1{,}000)\,[\mathcal{T}(\sqrt{(D/(2S-1))})].$$

For the term in brackets, the tail probability which is 4.8 standard deviations from the mean is 1/1,259,000 which gives a probability of error of about 1.6%.

For 4.8 standard deviations, we need **D** large enough to satisfy

$$4.8 = Z = \sqrt{(D/(2S-1))}, \text{ or } D = 4.8^2 * 39 = 898.56.$$

Thus:

- 98.4% of the time, a vector dimension of **D** = 899 will give error-free discrimination between 20 vectors in the sum and 1,000 additional random vectors.

- Similarly, we can consider a "Medium" sized system with up to 100 vectors summed together, and with 100,000 other random vectors. Here a 5.9 standard deviation value for D is required for 1.8% probability of error, which works out to a bound on required vector dimension of **D** = 6,927. (Details omitted.)

- Finally, we can consider a "Large" sized system with up to 1,000 vectors summed together, and with 1,000,000 other random vectors. Here a 6.7 standard deviation value for D is required for 1% probability of error, which works out to a bound on required vector dimension of **D** = 90,000.

By comparison, Plate gives simulation results [Figure 56] that for 99% error free performance in a small system with 14 bundled vectors and 1,000 additional random vectors, the required dimension **D** is between 850 and 900, which is comparable to our simulations with +1/-1 components.

He also derives a bound on required dimensionality where vector components are normally distributed. Letting $q$ be the probability of error, Plate derives:

$$\mathbf{D} < 8(\mathbf{S+1}) \ln(\mathbf{N}/q)$$

provided

$$\mathbf{D} > 2(\mathbf{S+1}) / \pi.$$

These bounds are consistent with Propositions 1-3.

For “Small”, “Medium”, and “Large” systems, Plate’s bound for 1% probability of error yields required dimensions of 2,000; 13,000; and 148,000 respectively. Thus these bounds are not quite as tight as those derived above, or possibly systems with continuous vector components require greater dimensions than systems with binary components.